\documentclass[10pt]{article}

\usepackage[preprint]{tmlr}

\usepackage{microtype}
\usepackage{graphicx}
\usepackage{booktabs}
\usepackage{multirow}
\usepackage{tabularx}
\usepackage{amsmath}
\usepackage{amssymb}
\usepackage{mathtools}
\usepackage{amsthm}
\usepackage{algorithm}
\usepackage{algorithmic}
\usepackage{xcolor}
\usepackage{colortbl}
\usepackage{tikz}
\usetikzlibrary{positioning, arrows.meta, fit, backgrounds, decorations.pathreplacing, calligraphy}
\usepackage{hyperref}
\usepackage{url}
\usepackage[capitalize,noabbrev]{cleveref}

\theoremstyle{plain}

\theoremstyle{definition}

\theoremstyle{remark}

\definecolor{tancolor}{RGB}{187,130,90}
\definecolor{slateblue}{RGB}{85,104,154}
\definecolor{lm_purple}{RGB}{227,227,240}
\definecolor{lm_purple_low}{RGB}{240,240,248}

\title{Minimal-Intervention KV Retention\\
       via Set-Conditioned Diversity}

\author{%
  \name Libo Sun \email libo@auburn.edu \\
  \addr Department of Computer Science and Software Engineering\\
        Auburn University, Auburn, AL, USA
  \AND
  \name Po-wei Harn \email harnpowei@ncu.edu.tw \\
  \addr Department of Information Management\\
        National Central University, Taoyuan, Taiwan
  \AND
  \name Peixiong He \email pzh0029@auburn.edu \\
  \addr Department of Computer Science and Software Engineering\\
        Auburn University, Auburn, AL, USA
  \AND
  \name Xiao Qin \email xqin@auburn.edu \\
  \addr Department of Computer Science and Software Engineering\\
        Auburn University, Auburn, AL, USA
}

\def\month{MM}
\def\year{YYYY}
\def\openreview{\url{https://openreview.net/forum?id=XXXX}}

\begin{document}

\maketitle

\makeatletter
\let\AND\@undefined
\makeatother


\begin{abstract}
KV-cache compression at small budgets is a crowded design space
spanning cache representation, head-wise routing, compression
cadence, decoding behavior, and within-budget scoring. We study
seven mechanisms across these five families on long-form
mathematical reasoning (MATH-500~\citep{hendrycks2021math}) at
budgets $b \in \{64, 128\}$, under an evaluation standard that
tightened over the study and converged on matched mean cache with
$n \geq 200$ on two distilled-reasoning models (Qwen-7B and
Llama-8B variants of DeepSeek-R1-Distill~\citep{deepseek2025r1}).
All seven were rejected as catalogue directions, one on
screening-grade evidence. We then propose
$\alpha$, a one-function modification to the
TriAttention~\citep{mao2026triattention} retention scorer that
replaces argmax-top-$k$ with greedy facility-location-inspired
selection under a V-space redundancy penalty controlled by a
single weight $\lambda$. A pre-registered protocol tunes $\lambda$
on a frozen development split and confirms on a disjoint held-out
split; with $\lambda = 0.5$, $\alpha$ clears Bonferroni on two of
the four (model, budget) cells (Qwen $b{=}128$ and Llama $b{=}64$),
no cell is significantly negative, and the pre-registered
Branch~A triggers. The finding is asymmetric: the surviving
mechanism was among the smallest tested, but minimality alone did
not predict survival~--- two comparably small scoring
modifications were also rejected~--- so what distinguished
$\alpha$ was its set-conditioned selection rule, in which each
retention decision depends on the already-retained set, rather
than its size. The combined matched-memory, sympy-graded,
held-out confirmation protocol is the evidence standard that made
the asymmetry visible.
\end{abstract}


\section{Introduction}
\label{sec:intro}

The KV cache is the dominant memory cost of long-context language-model
inference. At decode time, cached key and value tensors occupy memory
linear in sequence length, layers, heads, and head dimension, and grow
monotonically with generation. KV-cache compression---selecting a
budget-bounded subset of cached tokens at each compression
event~\citep{li2024snapkv,zhang2023h2o,liu2023scissorhands,xiao2024efficient}---trades
a small amount of accuracy for substantial memory savings. Below a
threshold, however, compression begins to degrade reasoning accuracy
visibly: at small budgets ($b \in \{64, 128\}$) on long-form reasoning
workloads, published mechanisms cluster within a few accuracy points
of one another, and each proposes a different intervention to push
the frontier.

\begin{figure}[t]
  \centering
  \begin{tikzpicture}[
    font=\footnotesize,
    every node/.style={inner sep=2pt},
    family/.style={
      rectangle, draw=black!75, line width=0.4pt,
      minimum width=2.05cm, minimum height=1.05cm,
      align=center, rounded corners=2.5pt,
      inner sep=3pt,
    },
    repbox/.style ={family, fill=blue!8!white},
    ctrlbox/.style={family, fill=yellow!12!white},
    sigbox/.style ={family, fill=red!10!white,
                    line width=0.65pt, draw=red!55!black},
    arr/.style={-{Latex[length=1.4mm, width=1.2mm]},
                line width=0.5pt, draw=black!70},
    axislabel/.style={font=\scriptsize\itshape, color=black!55},
    bracket/.style={line width=0.5pt, draw=black!55},
    surv/.style={font=\footnotesize, color=red!55!black},
  ]
    \node[repbox]  (state)    at (0.0, 0)    {\textbf{State}\\[1pt]\scriptsize\itshape what the cache holds};
    \node[repbox]  (routing)  at (3.1, 0)    {\textbf{Routing}\\[1pt]\scriptsize\itshape which heads see what};
    \node[ctrlbox] (cadence)  at (5.95, 0)   {\textbf{Cadence}\\[1pt]\scriptsize\itshape when to compress};
    \node[ctrlbox] (decoding) at (8.8, 0)    {\textbf{Decoding}\\[1pt]\scriptsize\itshape what the model emits};
    \node[sigbox]  (scoring)  at (11.65, 0)  {\textbf{Scoring}\\[1pt]\scriptsize\itshape which tokens kept};

    \draw[arr] (state)    -- (routing);
    \draw[arr] (routing)  -- (cadence);
    \draw[arr] (cadence)  -- (decoding);
    \draw[arr] (decoding) -- (scoring);

    \draw[bracket, decorate, decoration={brace, amplitude=4pt, raise=2pt}]
      (cadence.north west) -- (decoding.north east)
      node[midway, above=7pt, axislabel] {control-side};

    \node[axislabel, anchor=north] at ([yshift=-3pt]state.south)    {representation};
    \node[axislabel, anchor=north] at ([yshift=-3pt]routing.south)  {selection};
    \node[axislabel, anchor=north] at ([yshift=-3pt]scoring.south)  {selection};

    \node[surv, anchor=north]
      (alpha) at ([yshift=-22pt]scoring.south)
      {$\boldsymbol{\alpha}$ \textnormal{at} $\lambda{=}0.5$ \textnormal{---} \textbf{survived held-out confirmation}};
    \draw[-{Latex[length=1.6mm, width=1.4mm]}, line width=0.6pt,
          color=red!55!black]
      (alpha.north) -- ([yshift=-1pt]scoring.south);
  \end{tikzpicture}
  \caption{KV-retention pipeline at decode time: the five
    intervention surfaces examined in this study. State, Routing,
    and Scoring modify representation and selection over the cache;
    Cadence and Decoding are control-side interventions sharing a
    super-bracket. Seven mechanisms across these surfaces were
    tested and all rejected; the
    set-conditioned scoring intervention $\alpha$ at $\lambda{=}0.5$
    (marked) is the only survivor. Family-by-family verdicts are
    in Table~\ref{tab:verdicts} (Section~\ref{sec:designspace}).}
  \label{fig:pipeline}
\end{figure}
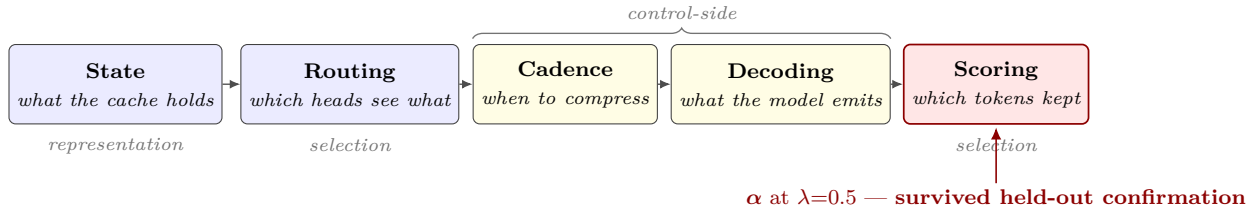

This paper asks a narrower question. Holding the model, budget,
routing policy, compression cadence, and decoding loop fixed,
\emph{which axis of intervention actually moves accuracy at matched
memory?} We test seven mechanisms across five families: \textbf{state}
(which K/V representations are stored), \textbf{routing} (which heads
or layers see which subset of the cache), \textbf{cadence} (when to
compress and with what budget across decode steps), \textbf{decoding}
(how the model emits text given that the cache is being compressed),
and \textbf{scoring} (which existing tokens are kept under a fixed
budget). The seven mechanisms are organized into a retrospective
design-space catalogue (Section~\ref{sec:designspace}). They were
gated sequentially, under a methodological standard that itself
tightened over the course of the study and converged on the one
later used for $\alpha$: matched mean cache instrumented during
decode, sympy-based grading on MATH-500~\citep{hendrycks2021math},
and $n \geq 200$ evaluation. Earlier gates were decided under that
standard as it then existed; Section~\ref{sec:designspace} flags
which kills rest on screening-grade ($n{=}50$ or single-model)
evidence. We then propose $\alpha$, a small, set-conditioned
scoring-axis modification, and evaluate it under a pre-registered protocol with
a frozen development split for hyperparameter selection and a
disjoint held-out split for confirmation. The seven catalogue
mechanisms were rejected; $\alpha$ survived. Figure~\ref{fig:pipeline}
sketches the five-axis pipeline and marks where $\alpha$ sits.

\paragraph{Contributions.}
The design-space study (Section~\ref{sec:designspace})
catalogues seven mechanisms across the five families above, gives
a falsified-mechanism verdict for each, and extracts five failure
modes that recur across families. $\alpha$
(Section~\ref{sec:alpha}) is a one-function modification to the
TriAttention~\citep{mao2026triattention} retention scorer that
replaces argmax-top-$k$ with a greedy facility-location-inspired
selector under a V-space redundancy penalty controlled by a
single weight $\lambda$; at $\lambda = 0$ the selector reduces
bitwise to top-$k$. The pre-registered protocol
(Section~\ref{sec:alpha_protocol}) is designed against these five
failure modes: a frozen MD5-bucketed dev/confirm split, a
$\lambda$-probe on the development split alone, and a
Bonferroni-corrected joint test on the held-out confirmation
split, with the decision rule and branch table committed in code
before the confirmation experiment runs. The held-out result
(Section~\ref{sec:alpha_results}) is that on MATH-500 with Qwen-7B
and Llama-8B variants of DeepSeek-R1-Distill~\citep{deepseek2025r1}
at $b \in \{64, 128\}$, $\alpha$ clears Bonferroni on two of the
four (model, budget) cells with no significantly negative cells
and triggers the pre-registered Branch~A; a post-confirmation
head-to-head against a SnapKV-style windowed-attention
baseline~\citep{li2024snapkv} reproduces the same shape. Beyond
the result, the positioning claim (Section~\ref{sec:lesson}) is
that in this regime the scoring axis was the only intervention
family to yield a survivor; minimality alone did not predict
survival, since two comparably small scoring modifications were
also rejected, and what set $\alpha$ apart was its set-conditioned
selection rule~--- a property of the design space distinct from
the accuracy headline itself.

\paragraph{Scope.}

The result is intentionally narrow. Our evaluation covers long-form
mathematical reasoning with two distilled-reasoning models in the
7--8B-parameter range. We do not claim transfer to other workloads
(LongBench~\citep{bai2024longbench}, code generation, multi-document
QA), to larger models, or to non-reasoning models without explicit
re-evaluation. The pre-registered protocol covers the method only; the
design-space study (Section~\ref{sec:designspace}) is retrospective
and is presented as such throughout.

\paragraph{Why this regime is worth isolating.}

Most published KV-compression methods report accuracy curves at
budgets where the baseline retention pipeline already handles the
workload well. The pressure point is the smallest budgets that remain
coherent at all---where every byte of cache is contested and small
intervention surfaces compete for the same accuracy budget. Long-form
mathematical reasoning is particularly demanding there because
chain-of-thought generation~\citep{wei2022cot} produces many
decode-time steps, so compression triggers fire repeatedly and
per-trigger errors accumulate; and because its closed-form answers
admit sympy-based grading, which removes a formatting-noise floor
that would otherwise dominate small-budget evaluation
(Section~\ref{sec:related} details the grading protocol).

\section{Related Work}
\label{sec:related}

We organize prior work along the five axes introduced in
Section~\ref{sec:intro} and conclude with the workload and
evaluation methodology that distinguish our protocol from the
published norm.

\paragraph{Scoring-family compression.}
The dominant published line of KV-cache compression operates on
the scoring axis: given a per-token retention score, retain the
top-$k$. H2O~\citep{zhang2023h2o} introduces the heavy-hitter
hypothesis~--- a small set of attention-heavy tokens carries most
of the model's future predictions and should be preserved.
SnapKV~\citep{li2024snapkv} uses end-of-prefill windowed attention
to identify which tokens future queries are likely to attend to
and then retains those tokens through generation.
Scissorhands~\citep{liu2023scissorhands} reframes retention
selection as a persistence-of-importance test, keeping tokens that
score high across multiple recent attention windows.
StreamingLLM~\citep{xiao2024efficient} keeps a small set of initial
``attention sink'' tokens alongside a recent window, exploiting
the empirical observation that early tokens receive
disproportionate attention mass independent of semantic
centrality. Quest~\citep{tang2024quest} introduces query-aware
sparsity, tracking per-page min/max keys and estimating page
criticality from the current query rather than precomputed
statistics. TriAttention~\citep{mao2026triattention}, our internal
baseline scorer (``$1d$'' throughout), uses precomputed Q/K
frequency statistics as a proxy for which positions future queries
will need. These methods all modify \emph{which} tokens are kept
under a fixed budget; $\alpha$ (Section~\ref{sec:alpha}) inherits
this scope, adding a single V-space diversity term and replacing
the modular top-$k$ argmax with a greedy
facility-location-inspired selector that conditions each pick on
the already-selected set. Closely related, R-KV~\citep{cai2025rkv}
identifies redundancy in K-space for reasoning models; we share
the redundancy-as-signal framing but operate in V-space and
through the existing retention scorer rather than as a separate
stage. The broader family of diverse subset selection via
determinantal point processes~\citep{kulesza2012dpp} offers an
alternative formalisation; we use a lighter facility-location
heuristic to stay one function away from the existing scorer.

\paragraph{Budget-allocation compression.}
A second line of work modifies how a global cache budget is split
across the model's structure. PyramidKV~\citep{cai2024pyramidkv}
allocates KV-cache budget per layer in a pyramidal shape,
motivated by the empirical observation that lower layers attend
more broadly than upper layers. Ada-KV~\citep{feng2025adakv}
allocates budget per attention head adaptively, moving budget
toward heads with higher retention need, and is intended to
compose with an underlying attention-based scorer (e.g., SnapKV)
rather than replace it. These methods are budget allocators, not
scoring functions: they operate on the same
retention/allocation/scoring axis as $\alpha$ but at a different
lever, and are in principle stackable with $\alpha$ as the
within-budget scorer; Section~\ref{sec:lesson_future} discusses
such compositions.

\paragraph{Quantization-based compression.}
Orthogonal to selection-based compression, several methods
compress the K and V tensors themselves via quantization:
KIVI~\citep{liu2024kivi} (asymmetric per-channel K and per-token V
quantization at 2 bits), KVQuant~\citep{hooper2024kvquant}
(non-uniform datatypes with per-channel scales),
KVTuner~\citep{li2025kvtuner} (mixed-precision),
MiniKV~\citep{yang2024minikv} (rank-based),
QAQ~\citep{dong2024qaq}, and ASVD~\citep{yuan2024asvd}.
\citet{brandon2024reducing} reduce KV-cache memory by sharing
the cache across layers. These directions are complementary to
scoring-axis interventions: quantization modifies how each
retained slot is represented, while selection modifies which
slots are retained. Our negative gate
\texttt{subspace\_moekv} (Section~\ref{sec:designspace}) is in
this family in spirit~--- per-head V is compressed via a low-rank
PCA basis~--- and was killed under our matched-memory protocol.

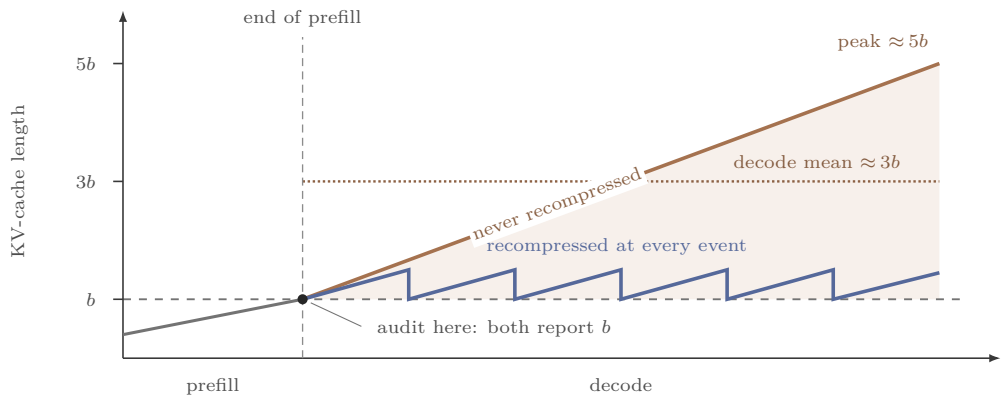
\begin{figure}[t]
  \centering
  \begin{tikzpicture}[
    x=1.08cm, y=0.78cm,
    font=\footnotesize,
    ax/.style={line width=0.6pt, draw=black!78},
    budgetline/.style={dashed, line width=0.7pt, draw=black!55},
    meanline/.style={densely dotted, line width=0.9pt, draw=tancolor!75!black},
    pre/.style={line width=1.15pt, draw=black!55},
    recomp/.style={line width=1.35pt, draw=slateblue},
    drift/.style={line width=1.35pt, draw=tancolor!88!black},
    lead/.style={line width=0.4pt, draw=black!55},
    slab/.style={font=\scriptsize, text=black!70},
  ]
    \fill[tancolor!12] (2.2,1) -- (10,5) -- (10,1) -- cycle;
    \draw[budgetline] (0,1)   -- (10.25,1);
    \draw[meanline]   (2.2,3) -- (10,3);
    \draw[ax,-{Latex[length=1.5mm]}] (0,0) -- (0,5.9);
    \draw[ax,-{Latex[length=1.5mm]}] (0,0) -- (10.75,0);
    \foreach \yy/\lab in {1/$b$, 3/$3b$, 5/$5b$}{%
      \draw[ax] (-0.12,\yy) -- (0,\yy);
      \node[slab, anchor=east] at (-0.22,\yy) {\lab};
    }
    \node[slab, rotate=90, anchor=south] at (-1.05,3) {KV-cache length};
    \draw[densely dashed, line width=0.5pt, draw=black!45]
      (2.2,0) -- (2.2,5.45);
    \node[slab, anchor=south] at (2.2,5.47) {end of prefill};
    \node[slab, anchor=north] at (1.1,-0.18) {prefill};
    \node[slab, anchor=north] at (6.1,-0.18) {decode};
    \draw[pre] (0,0.4) -- (2.2,1);
    \draw[drift] (2.2,1) -- (10,5)
      node[pos=0.40, sloped, fill=white, inner sep=1.2pt, slab,
           text=tancolor!72!black]
      {never recompressed};
    \draw[recomp] (2.2,1) -- (3.5,1.5) -- (3.5,1) -- (4.8,1.5) -- (4.8,1)
       -- (6.1,1.5) -- (6.1,1) -- (7.4,1.5) -- (7.4,1)
       -- (8.7,1.5) -- (8.7,1) -- (10,1.45);
    \node[slab, anchor=south, text=slateblue] at (6.05,1.6)
      {recompressed at every event};
    \node[slab, anchor=south, text=tancolor!70!black] at (8.5,3.07)
      {decode mean ${\approx}\,3b$};
    \node[slab, anchor=south east, text=tancolor!70!black] at (9.98,5.06)
      {peak ${\approx}\,5b$};
    \fill[black!85] (2.2,1) circle (1.9pt);
    \draw[lead] (2.3,0.94) -- (2.92,0.54);
    \node[slab, anchor=west] at (3.0,0.5)
      {audit here: both report~$b$};
  \end{tikzpicture}
  \caption{Why end-of-prefill memory matching is insufficient
    (schematic; not measured cache trajectories). Two methods
    identical at the end of prefill diverge
    sharply during decode, so the protocol matches \emph{mean} cache
    over the full generation, not end-of-prefill cache
    (Section~\ref{sec:failure_modes}).}
  \label{fig:matchedmem}
\end{figure}

\paragraph{Workload and evaluation methodology.}
\label{sec:related_eval}
We evaluate on long-form mathematical reasoning generated by
distilled-reasoning models~\citep{deepseek2025r1}, which distill
from the Llama 3~\citep{dubey2024llama3} and Qwen base families and
trace their math capability to math-specialized lines such as
DeepSeek-Math~\citep{shao2024deepseekmath}; chain-of-thought
prompting~\citep{wei2022cot} unlocks multi-step reasoning at the
cost of long decode horizons, the regime where KV-cache compression
matters most. The MATH benchmark~\citep{hendrycks2021math} and
GSM8K~\citep{cobbe2021gsm8k} provide closed-form answers that admit
sympy-based grading, which normalizes equivalent answer forms
(\texttt{\textbackslash frac} vs.\ \texttt{\textbackslash dfrac},
simplification, sign convention) and removes the 5--10\,pp
formatting-noise swings that exact-string graders show at small
$n$; at $b = 64$, $44$--$62$\,\% of incorrect runs produce no
parseable boxed answer at all, which an exact-string grader would
conflate with reasoning failure. Most published KV-compression
methods report on LongBench~\citep{bai2024longbench},
RULER~\citep{hsieh2024ruler}, or similar retrieval benchmarks
rather than long-form generation accuracy. Two protocol choices
distinguish ours. End-of-prefill memory matching is insufficient
when the decode horizon is long~--- methods that do not recompress
during decode use multiples of the nominal budget
(Figure~\ref{fig:matchedmem})~--- so we match \emph{mean} cache
instrumented over the full generation; several negative gates were
green on end-of-prefill matching and red under it. Separately,
reporting the same dataset for tuning and headline inflates apparent
performance~\citep{bouthillier2021variance}, so we separate the two
via a frozen dev/confirm split with the decision rule committed in
code before confirmation, and argue pre-registration should be
standard here.

%

\section{Design-Space Study: Why Most Mechanisms Failed}
\label{sec:designspace}

This section is a retrospective design-space study, not a
pre-registered sweep. We tested seven mechanisms across the
KV-retention pipeline; every structural intervention failed under
matched-memory evaluation, and the surviving design is a small
scoring fix (Section~\ref{sec:alpha}). The negatives are what
make the survivor interpretable: the contribution is not that one
mechanism worked, but that the lone survivor is a scoring-axis
mechanism of a particular form (Section~\ref{sec:bridge}) and that
the failures fall along mechanism-class boundaries rather than
implementation-detail boundaries.

At each decode step the model holds a key--value cache of length
up to some budget $B$. A KV-retention pipeline must answer five
questions, in roughly this order. \emph{State} asks what the
cache holds: per-token K/V tensors by default, with alternatives
that compress or replace tensors via PCA bases, span-mean
``synthetic'' summaries, or low-rank projections. \emph{Routing}
asks which heads or layers see which subset of the cache: a
single global policy by default, with alternatives that admit
per-head or per-layer choices among multiple scoring strategies.
\emph{Cadence} asks when compression is triggered and with what
budget across decode steps: a fixed budget at every trigger by
default, with alternatives that modulate budget by step index.
\emph{Decoding} asks what the model emits while the cache is
being compressed: unmodified generation by default, with
alternatives that prompt for dense anchor tokens the eviction
scorer can pin. \emph{Scoring} asks which tokens are kept under
fixed budget, routing, cadence, and decoding behavior: this is
where attention-based heuristics
(TriAttention~\citep{mao2026triattention},
SnapKV~\citep{li2024snapkv}, H2O~\citep{zhang2023h2o}) and our
$\alpha$ regularizer live. The five families fall along three
axes: state modifies the \emph{representation} of cached entries,
routing and scoring modify the \emph{selection} over cached
entries, and cadence and decoding are control-side~--- they
modify when to compress and what the model emits without changing
what the cache holds at any given moment. Figure~\ref{fig:pipeline}
(page~\pageref{fig:pipeline}) shows the five surfaces with cadence
and decoding under a shared ``control-side'' super-bracket; in the
verdict table below they are tracked separately because they were
independent hypotheses with independent kills.

\subsection{Verdict Table}
\label{sec:verdicts}

Table~\ref{tab:verdicts} summarizes every mechanism we tested,
organized by family. Each of the seven negative gates corresponds to
a single entry in the negative-results catalog distributed with this
paper; the shaded bottom row is the surviving mechanism $\alpha$.
The synthetic-memory state family explored four sub-attempts before we
closed it at family level; for table compactness those four are
reported as a single row with a footnote pointer.

\begin{table}[!ht]
\centering
\small
\setlength{\tabcolsep}{4pt}
\begin{tabularx}{\linewidth}{@{}llXlp{2.4cm}@{}}
\toprule
\textbf{Family} & \textbf{Gate} & \textbf{Hypothesis tested}
  & \textbf{Verdict} & \textbf{Failure mode} \\
\midrule
\multirow{2}{*}{State}
  & \texttt{subspace}
  & Per-head $V$ is low-rank; a rank-32 PCA basis holds quality at ${\sim}1.6\times$ memory saving.
  & Killed
  & Proxy-to-runtime drift \\
  & \texttt{syn\_mem}\textsuperscript{$\dagger$}
  & Span-mean ``synthetic'' slots plus a $\log(N)$ mass bias beat per-token retention at extreme budgets.
  & Killed
  & Family-level bundling \\
\midrule
Routing
  & \texttt{static\_route}
  & An offline-calibrated per-head choice between uniform and attention scoring beats either alone.
  & Killed
  & Proxy-to-runtime drift \\
\midrule
Cadence
  & \texttt{schedule}
  & Round-indexed budget schedules favor high-risk reasoning stretches at matched mean cache.
  & Killed
  & Iso-cache underperformance \\
\midrule
Decoding
  & \texttt{state\_emit}
  & Prompted compact ``State:'' anchor lines give the eviction scorer dense pinnable tokens.
  & Killed
  & Iso-cache underperformance \\
\midrule
\multirow{3}{*}{Scoring}
  & \texttt{variant\_b}
  & A mean-$V$ cosine prior ($+\,\lambda\cos(V,\text{dir})$) adds signal beyond attention-only scoring.
  & Killed
  & Proxy-to-runtime drift \\
  & \texttt{Q\_stats}
  & Recalibrating TriAttention's Q/K statistics on math traces improves \texttt{1d} at aggressive budgets.
  & Killed\textsuperscript{$\ddagger$}
  & $n{=}50$ instability \\
  \cmidrule(l){2-5}
  & \cellcolor{gray!12}$\boldsymbol\alpha$ \textbf{($\lambda{=}0.5$)}
  & \cellcolor{gray!12}Set-conditioned greedy selection with a cosine-against-retained $V$-diversity penalty.
  & \cellcolor{gray!12}\textbf{Survived}
  & \cellcolor{gray!12}n/a \\
\bottomrule
\end{tabularx}
\caption{Verdict summary across mechanism families. The bottom row,
shaded, is the surviving mechanism examined in detail in
Sections~\ref{sec:alpha}--\ref{sec:alpha_results}.
\textsuperscript{$\dagger$}\,The
synthetic-memory state family closed at family level after four
consecutive sub-attempts (span-mean prefill states, bounded variant,
and two static/dynamic \texttt{sel\_hybrid} formulations). All four
failed iso-cache evaluation; we report them as a single family-level
kill in the table and itemize them in the catalog.
\textsuperscript{$\ddagger$}\,The \texttt{Q\_stats} gate was decided
on an $n{=}50$, single-model (Qwen-7B) screening evaluation that
predates the matched-memory, $n{\geq}200$, two-model standard later
applied to the other catalog gates and to $\alpha$; we report it as
a screening-grade kill (Section~\ref{sec:failure_modes}).}
\label{tab:verdicts}
\end{table}

The coverage is asymmetric, and we flag it. Across the four
structural families we ran five catalog mechanisms (and roughly ten
sub-attempts once the synthetic-memory family is itemized), against
three scoring-family mechanisms (variant-B V-direction prior,
reasoning-statistics recalibration, and $\alpha$). Scoring received
the same or more design effort per mechanism; the family is narrower
because, once budget, routing, cadence, and decoding are fixed, the
remaining design surface is the score function itself, which admits
fewer qualitatively distinct interventions than the structural
surfaces it sits on top of. Section~\ref{sec:bridge} makes this
argument precise.

\subsection{Shared Failure Modes}
\label{sec:failure_modes}

The kill verdicts in Table~\ref{tab:verdicts} are not independent.
Five failure modes recur across the mechanism families. Four of them
carry the kills in Table~\ref{tab:verdicts}~--- its \textbf{Failure
mode} column names, for each gate, the mode its kill rests on~--- and
the fifth, cross-model transfer failure, is a validity threat that no
single gate's kill rests on but that the protocol of
Section~\ref{sec:alpha_protocol} is built to defend against. One
further effect, $b{=}64$ extraction collapse, is an evaluation-regime
caveat rather than a mechanism-failure mode; we describe it last.

The first mode, \emph{iso-cache underperformance}, has the widest
reach. Once mean cache length is matched at decode time --- not merely
at the end of prefill --- structural mechanisms sit below the linear
interpolation between fixed-budget reference points. End-of-prefill
matching is not matched memory: a mechanism that does not recompress
during decode can silently use 5$\times$ more cache than its reported
budget would suggest (Figure~\ref{fig:matchedmem}), so we require
per-problem peak/mean cache instrumentation as a gate-passing
condition. Two gates fail this way through gate-specific symptoms. The
cadence schedules (\texttt{schedule}) sat $\sim$6\,pp below the
iso-cache interpolation whether budget was front-loaded or
back-loaded, and a budget-step inversion test confirmed the shortfall
is phase-independent; the decoding gate (\texttt{state\_emit}) instead
saw its compact ``State:'' anchor lines crowd reasoning content out of
the cache at $b{=}128$. The local symptom differs, but the mode is the
same.

Two further modes share both a cause and a remedy: each is evidence
read off a shortcut that a proper test does not bear out, and each is
therefore treated as a screening signal rather than a verdict. Under
\emph{$n{=}50$ instability}, headline deltas at $n{=}50$ on MATH-500
routinely shrank substantially at $n{=}200$: an early $\alpha$-line
measurement at $n{=}50$ on Qwen-7B reported a mean lift of
$+5.6$\,pp at aggressive budgets, yet the same mechanism at $n{=}200$
on two models shrank to between $+0.2$ and $+0.5$\,pp, and the
reasoning-statistics gate was even sign-inconsistent across budgets at
fixed $n{=}50$ ($+6$\,pp at $b{=}128$, $-2$\,pp at $b{=}64$). We
therefore treat any $n{=}50$ decision as screening only and require
$n \geq 200$ to confirm; because the \texttt{Q\_stats} kill rests on
exactly this $n{=}50$, single-model evidence and was not re-run at the
$n \geq 200$ two-model standard, Table~\ref{tab:verdicts} marks it as
a screening-grade kill rather than a fully protocol-matched one.
\emph{Proxy-to-runtime drift} is the same hazard in metric form:
pre-RoPE Q/K attention surrogates, oracle-keep capture rates, and
mean-V cosine similarities all failed to predict matched-memory task
accuracy, and the variant-B, subspace-moeKV, and static-route gates
were each green on a proxy and red at runtime. In both cases the
decision rule is runtime accuracy on the matched-memory protocol; a
proxy or a small-$n$ delta only screens.

A fourth mode, \emph{family-level bundling}, is a discipline on
negative inference: a single sub-attempt's failure is not a class
kill. The synthetic-memory state family was rejected only after four
consecutive sub-attempts --- span-mean prefill, a bounded re-gate, and
static and dynamic \texttt{sel\_hybrid} --- had all failed iso-cache.
That family-level kill is load-bearing: it is what licenses the claim
``state-side rewrites do not survive matched-memory evaluation''
rather than the weaker ``this particular state-side rewrite did not.''

The fifth mode is the one no gate's kill rests on.
\emph{Cross-model transfer failure} is a validity threat the protocol
must defend against regardless: interventions calibrated on Qwen-7B
did not always survive transfer to Llama-8B at $n \geq 200$. The
$\alpha$-only mechanism at $\lambda{=}1$ --- an earlier checkpoint on
the path to the Branch-A result --- produced a Qwen-7B lift around
$+1.7$\,pp at $p \approx 0.05$ on the $b{=}128$ cell while leaving the
Llama-8B $b{=}64$ cell statistically null ($p = 0.92$); only after
re-tuning $\lambda$ to $0.5$ on a held-out dev split did Llama recover
a Bonferroni-significant lift. Several other gates evaluated on Qwen
alone --- the reasoning-statistics recalibration among them ---
provide no Llama-side evidence either way, which is itself a
transfer-failure mode: a single-model gate is not gate-passing. We
therefore require both Qwen-7B and Llama-8B in any gate-passing
evaluation, and flag single-model wins as such.

One further effect is a caveat rather than a failure mode, and it
bounds how the $b{=}64$ cells should be read. At budgets near 64 on
long-form math reasoning, 44--62\,\% of incorrect runs fail with
\texttt{extracted\_answer = None} --- the model produces no parseable
boxed answer at all, regardless of reasoning quality. This $b{=}64$
extraction collapse afflicts every method, the baseline included, so
it kills no gate; but it does mean that aggressive-budget evaluation
at $b{=}64$ partly tests structured-output fluency rather than
retention quality, so we report $b{=}64$ alongside $b{=}128$ to keep
both signals visible and flag the affected cells where they arise
(Section~\ref{sec:alpha_honest}).

These five failure modes and the $b{=}64$ caveat are the
methodological backbone of the evaluation
in Sections~\ref{sec:alpha}--\ref{sec:alpha_results}.
The protocol was designed around these risks: a frozen dev/confirm
split (against $n{=}50$ instability and threshold chasing);
evaluation on both Qwen-7B and Llama-8B (against cross-model
transfer failure); per-problem mean cache instrumented during
decode (against iso-cache underperformance); runtime accuracy as
the decision variable rather than any proxy metric (against
proxy-to-runtime drift); and a
Bonferroni-corrected joint test across the four (model, budget)
cells (against per-cell multiple-comparison artifacts).

\subsection{What Distinguished the Survivor: Preview, with Caveats}
\label{sec:bridge}

The pattern in Table~\ref{tab:verdicts} is the design-space claim:
the sole survivor is on the scoring axis. Structural mechanisms
change what the cache contains, what routes to it, when it is
compressed, or what gets generated~--- each family adds degrees of
freedom that have to be paid for in retention quality at matched
memory, and the payments were not recovered. The $\alpha$ selector
(Section~\ref{sec:alpha}) operates on the same K/V tensors, routing,
budget schedule, and generated text as the baseline, adding a single
diversity term to the existing $1d$ score; at $\lambda = 0$ it is a
bitwise no-op against the baseline.

A small intervention surface is, on its own, not what separated the
survivor from the failures. The \texttt{variant\_b} gate is also a
one-term, one-weight scoring modification~--- it adds
$\lambda \cos(V, \text{dir})$ to the score~--- and it was rejected;
so was \texttt{Q\_stats}, a recalibration that adds no term at all.
Of the three scoring-family mechanisms we tested, two failed, and the
intervention surface does not separate them: \texttt{variant\_b}'s is
if anything smaller than $\alpha$'s, since $\alpha$ additionally
replaces top-$k$ with a greedy selector. That extra machinery is not
a free design choice~--- it is \emph{forced} by what does distinguish
$\alpha$: its selection rule is \emph{set-conditioned}. The redundancy
penalty in Equation~\ref{eq:marginal} is evaluated against the
already-retained set, so each pick depends on the picks before it,
which an independent per-token argmax cannot express.
\texttt{variant\_b}'s cosine-to-a-fixed-direction prior and
\texttt{Q\_stats}'s recalibrated statistics both rescore tokens
\emph{modularly}~--- token by token, with no dependence on the
retained set~--- leaving top-$k$ intact. The design-space reading is
therefore two-part: the scoring axis was the only family to yield a
survivor, and the feature separating that survivor from the two
scoring-family failures was set-conditioned redundancy control, not
minimality as such. This is a single-survivor observation rather than
a controlled comparison; Section~\ref{sec:lesson_stated} states it as
a design lesson with that caveat, and Section~\ref{sec:alpha_results}
provides the matched-memory evidence under the pre-registered
protocol.

Two caveats bound how Table~\ref{tab:verdicts} should be read. The
seven negative gates were sequential hypothesis tests with shifting
motivations, not a pre-designed factorial sweep across the family
taxonomy: the taxonomy is the contribution of this section, the
chronology is a fact about how we arrived at it, and the catalog
distributed with this paper records the full timeline. Separately,
the pre-registered protocol applies to $\alpha$
(Sections~\ref{sec:alpha}--\ref{sec:alpha_results}), not to the
negative gates that pre-date it: Section~\ref{sec:designspace} is
retrospective, and the strength of the design-space claim rests on
the confirmation in Section~\ref{sec:alpha_results}, not on the
negatives alone.

\section{$\alpha$: Method and Protocol}
\label{sec:alpha}

This section defines $\alpha$ and the pre-registered protocol used to
evaluate it. $\alpha$ is a one-function modification to the
TriAttention~$1d$ retention scorer~\citep{mao2026triattention} that
replaces argmax-top-$k$ with greedy facility-location-inspired
selection under a V-space redundancy penalty. The companion protocol
(Section~\ref{sec:alpha_protocol}) is designed against the failure
modes catalogued in Section~\ref{sec:designspace}: a frozen
$201$-problem development split for $\lambda$ tuning and a disjoint
$299$-problem held-out split for confirmation, with three seeds,
cluster bootstrap, and a Bonferroni-corrected joint test across four
(model, budget) cells. Section~\ref{sec:alpha_results} reports
Phase~1 and Phase~2 results and reads the verdict against a
pre-committed branch table.

\subsection{Method: A V-Space Diversity Penalty}
\label{sec:alpha_method}

Let $s \in \mathbb{R}^n$ be the per-token retention scores produced by
the TriAttention $1d$ scorer at a compression event, where $n$ is the
current cache length and $k \le n$ is the post-compression budget. The
baseline retention rule is modular top-$k$: keep the indices in
$\arg\max_k s$. We replace this with a greedy facility-location-style
selector under a V-space redundancy penalty.

For each token $i$, define a V-signature $v_i \in \mathbb{R}^d$ by
averaging the value tensor across layers and across attention heads
and unit-normalizing:
\begin{equation}
  v_i = \frac{1}{\|u_i\|_2 + \varepsilon} \, u_i,
  \quad
  u_i = \frac{1}{L H} \sum_{\ell=1}^{L} \sum_{h=1}^{H} V^{(\ell, h)}_i,
  \label{eq:signatures}
\end{equation}
where $V^{(\ell, h)}_i \in \mathbb{R}^{d_{\text{head}}}$ is the value
vector for token $i$ at layer $\ell$, head $h$, and $L, H$ are the
layer and head counts. Cosine similarity between two unit-normalized
signatures reduces to a dot product:
$\cos(v_i, v_j) = v_i^\top v_j$.

Given an already-selected subset $S_t$ of size $t$, the marginal gain
of adding candidate token $i \notin S_t$ is
\begin{equation}
  \Delta(i \mid S_t) \;=\; s_i \;-\; \lambda \,\max\!\Big(0,\ \max_{j \in S_t} \cos(v_i, v_j)\Big).
  \label{eq:marginal}
\end{equation}
The outer $\max(0, \cdot)$ floor reflects an implementation choice:
candidates anti-aligned with the selected set should not receive a
\emph{negative} redundancy penalty (i.e., a bonus). A token whose
V-signature is roughly orthogonal to or anti-aligned with $S_t$ is
treated as having zero redundancy with it. The budget is filled
greedily:
\begin{equation}
  i_{t+1} \;=\; \arg\max_{i \notin S_t} \Delta(i \mid S_t),
  \qquad t = 1, \ldots, k-1.
  \label{eq:greedy}
\end{equation}
The first pick $i_1 = \arg\max_i s_i$ is the modular argmax (matching
the baseline). At $\lambda = 0$ the selector reduces exactly to
top-$k$; this is enforced by an explicit fast path so that
$\lambda = 0$ is a bitwise no-op against the
TriAttention~\citep{mao2026triattention} baseline.

Algorithm~\ref{alg:alpha} summarizes the selector.

\begin{algorithm}[t]
\caption{$\alpha$ retention selection (one compression event).}
\label{alg:alpha}
\begin{algorithmic}[1]
\REQUIRE base scores $s \in \mathbb{R}^n$,
         V-signatures $V \in \mathbb{R}^{n \times d}$ with rows
         unit-normalized,
         budget $k$, regularization $\lambda \ge 0$.
\IF{$\lambda = 0$}
  \STATE \textbf{return} indices of $\text{top-}k(s)$ \quad \emph{(top-$k$ fast path)}
\ENDIF
\STATE $\text{Sims} \gets V V^\top$ \quad // $n \times n$ cosine matrix
\STATE $S \gets \emptyset$;\; $m \gets \mathbf{0}_n$
       \quad // $m_i = \max\!\big(0,\ \max_{j \in S} \cos(v_i, v_j)\big)$
\STATE $i_1 \gets \arg\max_i s_i$;\; $S \gets S \cup \{i_1\}$;\;
       $m \gets \max(m, \text{Sims}[i_1, :])$
\FOR{$t = 1, \ldots, k - 1$}
  \STATE $a \gets s - \lambda m$;\; $a[S] \gets -\infty$
  \STATE $i \gets \arg\max_j a_j$
  \STATE $S \gets S \cup \{i\}$;\;
         $m \gets \max(m, \text{Sims}[i, :])$
\ENDFOR
\STATE \textbf{return} $\text{sort}(S)$
\end{algorithmic}
\end{algorithm}

Equation~\ref{eq:greedy} is a facility-location-inspired greedy
selector~\citep{krause2014submodular}: each pick balances the
per-token base score against a redundancy term against the
already-selected set. Compared to the modular $\arg\max_k s$
baseline, the per-step decision now depends on the current
selection, breaking the indices-are-independent property of
top-$k$. Our objective is not in general submodular, so we do not
claim the $(1-1/e)$-style approximation guarantee for greedy
submodular maximization~\citep{nemhauser1978approx}; the empirical
value of the selector is established by the held-out evaluation
in Section~\ref{sec:alpha_phase2}.

The modification is intentionally minimal: the baseline K and V
tensors, routing, budget schedule, and decoding procedure are all
unchanged, and only the $k$-element selection step inside the
existing $1d$ scorer is modified. The hyperparameter $\lambda$
controls the strength of the diversity penalty and is tuned in
Phase~1. Algorithm~\ref{alg:alpha} computes an $n \times n$ cosine
matrix once per compression event and then runs $k - 1$ greedy
picks with constant-time marginal-gain updates, an
$O(n^2 d + n k)$ step over the candidate pool of size $n$; at
$b \in \{64, 128\}$ this overhead is negligible against the rest
of decode in our matched-memory harness, and we did not
specifically optimize it.

\subsection{Pre-registered Protocol}
\label{sec:alpha_protocol}

Section~\ref{sec:failure_modes} listed five failure modes that an
honest evaluation in this setting has to defend against; the protocol
below addresses each, and was registered as executable code before
Phase~2 ran. We hash each MATH-500~\citep{hendrycks2021math}
problem's \texttt{unique\_id} with MD5, bucket by
($\text{md5} \bmod 5$), and assign buckets $\{0, 1\}$ to a
201-problem development split and buckets $\{2, 3, 4\}$ to a
299-problem held-out confirmation split; the split is deterministic,
is produced by \texttt{alpha\_paper\_split.py}, and was committed
before Phase~1 rather than recomputed once Phase~1 results were
observed. The Phase~1 $\lambda$-selection
(\texttt{analyze\_alpha\_lambda\_probe.py}) and the Phase~2 cluster
bootstrap and Branch~A/B/C decision
(\texttt{analyze\_alpha\_paper\_confirmation.py}, whose
branch-decision function is the rule stated under \emph{Decision
rule} below) are committed code as well: the branch-decision code
carries development-repository commit \texttt{42b8e75} of
2026-04-30, six days before the Phase~2 confirmation run of
2026-05-06, and was not modified afterwards. All three scripts and
the frozen split files ship with the public release described in
Section~\ref{sec:lesson_future}, so the decision rule can be read
and the split reproduced.

\paragraph{Phase 1 ($\lambda$-probe; dev split only).}
Phase 1 is a lightweight $\lambda$-selection screen, not a full
factorial: we sweep $\lambda \in \{0.5, 1.0, 1.5\}$ across two models
(DeepSeek-R1-Distill-Qwen-7B, DeepSeek-R1-Distill-Llama-8B) and two
budgets ($b \in \{64, 128\}$) at a single seed, on the 201-problem
dev split only. The $\lambda = 1.0$ point reuses an existing
gate~(Appendix~\ref{app:phase1}); the $\lambda \in \{0.5, 1.5\}$
points required eight new evaluations. The $\lambda = 1.5$ arm is
incomplete on three of four cells
(Appendix~\ref{app:phase1} and Table~\ref{tab:phase1} report the
exact per-cell $n$); we disclose the gap, verify the ranking is
robust to it, and rely on Phase~2 to carry the central claim. The
decision rule is fixed in advance: pick the $\lambda$ that maximizes
the mean of $\Delta(\text{$1d_\text{div}$} - 1d)$ across the four
(model, budget) cells. Ties within $\pm 0.5$\,pp are broken toward
the smaller $\lambda$ (Occam).

\paragraph{Phase 2 (confirmation; confirm split only).}
The winning $\lambda$ from Phase~1 is evaluated on the 299-problem
confirm split with three seeds per cell, two models, two budgets~---
twelve cells total, all sympy-regraded. The 1d baseline is reused
from the prior gate (subsetted to the confirm IDs; no new compute).
We report per-cell paired deltas with cluster bootstrap on per-problem
accuracy differences ($n_\text{boot} = 10{,}000$).

\paragraph{Decision rule.}
The four (model, budget) cells form a Bonferroni family. Per-cell
significance uses $\alpha = 0.05$; joint significance after
Bonferroni correction uses $\alpha = 0.05/4 = 0.0125$. Branch
decisions were pre-committed before unblinding, mutually exclusive.
\emph{Branch~A} (primary-claim success) requires at least one
Qwen cell with $\Delta > 0$ at per-cell $p < 0.05$, a positive
Llama mean $\Delta$ across cells, and no significantly negative
Llama cell. \emph{Branch~B} (restricted generalization) satisfies the Qwen
condition but fails one of the two Llama conditions.
\emph{Branch~C} (no central $\alpha$ claim) has no qualifying Qwen
cell. One asymmetry in the rule as registered: the
no-significantly-negative guard is evaluated over the Llama cells
only, an oversight in the committed rule that we did not catch
before unblinding. We report against the rule as registered, and
note in Section~\ref{sec:alpha_branch} that the Phase~2 verdict is
unchanged under the stricter symmetric guard~--- no significantly
negative cell on \emph{either} model~--- which the data also satisfy.

\section{Confirmation Results}
\label{sec:alpha_results}

This section reports the held-out Phase~2 confirmation for $\alpha$
under the protocol of Section~\ref{sec:alpha_protocol}, reads the
verdict against the pre-committed branch table, and accounts for the
two non-significant cells.

Phase~1 swept $\lambda \in \{0.5, 1.0, 1.5\}$ on the 201-problem
development split and applied the pre-registered rule~--- maximize
the mean of $\Delta(\text{$1d_\text{div}$} - 1d)$ across the four
(model, budget) cells, ties broken toward the smaller $\lambda$. The
winner is $\lambda = 0.5$ (mean development-split
$\Delta = +0.75$\,pp, against $+0.43$ at $\lambda{=}1.5$ and $-0.50$
at $\lambda{=}1.0$); Appendix~\ref{app:phase1} reports the full
per-cell development table. Phase~2 evaluates the winning $\lambda$
only.

\subsection{Phase 2: Held-Out Confirmation}
\label{sec:alpha_phase2}

\begin{figure}[t]
  \centering
  \includegraphics[width=0.95\linewidth]{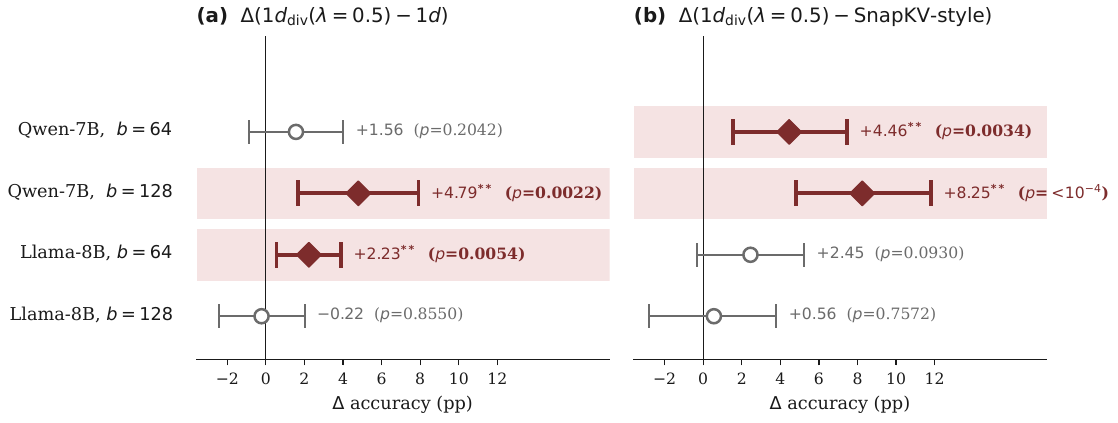}
  \caption{Side-by-side forest plots of two $\alpha$-vs-baseline
    contrasts on the same held-out confirm split. Three seeds,
    two-sided cluster bootstrap ($n_\text{boot} = 10{,}000$),
    Bonferroni-corrected jointly over the four (model, budget) cells
    ($\alpha = 0.0125$, marked \textbf{**}). Panel \textbf{(a)}:
    $\Delta(1d_\text{div}(\lambda{=}0.5) - 1d)$~--- the pre-registered
    Phase~2 confirmation. Panel \textbf{(b)}:
    $\Delta(1d_\text{div}(\lambda{=}0.5) - \text{SnapKV-style})$~---
    the post-confirmation head-to-head, reported in
    Appendix~\ref{app:snapkv}. Filled brick diamonds with
    shaded row tints mark Bonferroni-pass cells; open gray circles
    mark non-significant ones; inline annotations show the $\Delta$
    point estimate (pp) and two-sided $p$.}
  \label{fig:confirmation}
\end{figure}

Phase~2 evaluates $1d_\text{div}$ at $\lambda = 0.5$ against $1d$ on
the held-out confirm split, with three seeds per cell.
Table~\ref{tab:phase2} and Figure~\ref{fig:confirmation}\,(a) report
per-problem cluster-bootstrap estimates of
$\Delta(\text{$1d_\text{div}$} - 1d)$ with two-sided $p$-values and
95\,\% CIs.

\begin{table}[tbp]
\centering
\small
\begin{tabular}{@{}llcccc@{}}
\toprule
Model & $b$ & $\Delta$ (pp) & 95\,\% CI & $p$ & Bonferroni \\
\midrule
qwen7b  & 64  & $+1.56$  & $[-0.89, +4.01]$  & $0.20$    & --- \\
\textbf{qwen7b}  & \textbf{128} & $\boldsymbol{+4.79}$ & $\boldsymbol{[+1.67, +7.92]}$ & $\boldsymbol{0.0022}$ & \textbf{PASS} \\
\textbf{llama8b} & \textbf{64}  & $\boldsymbol{+2.23}$ & $\boldsymbol{[+0.56, +3.90]}$ & $\boldsymbol{0.0054}$ & \textbf{PASS} \\
llama8b & 128 & $-0.22$  & $[-2.45, +2.01]$  & $0.86$    & --- \\
\bottomrule
\end{tabular}
\caption{Phase 2 confirmation on the held-out split. Per-cell paired
delta with cluster bootstrap on per-problem accuracy differences
($n_\text{boot} = 10{,}000$). Joint Bonferroni threshold
$\alpha = 0.0125$. Two of four cells clear Bonferroni; no cell has a
significantly negative effect.}
\label{tab:phase2}
\end{table}

$\lambda = 0.5$ was selected on the disjoint development split by
the pre-registered rule, not on held-out confirm; on the confirm
split it strictly improves on the earlier $\lambda = 1.0$
checkpoint at all four cells, including flipping Llama $b{=}64$
from null to Bonferroni-pass (Appendix~\ref{app:phase1}). The split
structure makes that Llama $b{=}64$ pass especially clean: the cell
stood at $\Delta = +0.0$\,pp on the development split, so the
selection of $\lambda = 0.5$ drew no signal from it, and its
$+2.23$\,pp confirm result is free of dev-to-confirm selection
bias.

\label{sec:alpha_branch}
The pre-committed branch decision reads directly from
Table~\ref{tab:phase2}. Qwen has a Bonferroni-passing cell at
$b{=}128$, the Llama mean $\Delta$ across cells is $+1.00$\,pp
(positive), and the Llama $b{=}128$ null ($p = 0.86$) is well
within the null region rather than significantly negative. All
three Branch-A conditions hold; \textbf{Branch~A} triggers. The
verdict does not depend on the Llama-only asymmetry in the
registered guard (Section~\ref{sec:alpha_protocol}): under the
stricter symmetric form~--- no significantly negative cell on
either model~--- the two non-significant cells (Qwen $b{=}64$,
$p = 0.20$; Llama $b{=}128$, $p = 0.86$) sit within their null
regions, so no cell on either model is significantly negative and
Branch~A still triggers.

\subsection{Honest Negatives}
\label{sec:alpha_honest}

Two cells in Table~\ref{tab:phase2} did not clear per-cell
significance and we report them as such. Qwen $b{=}64$
($\Delta = +1.56$\,pp, $p = 0.20$) is positive in direction, like two of
the other three cells, but its 95\,\% CI
contains zero; the underlying baseline accuracy is only $\sim 8\%$
and roughly half of the failures at $b{=}64$ are extraction-collapse
rather than reasoning quality
(Section~\ref{sec:failure_modes}), so the effective signal is small.
Llama $b{=}128$ ($\Delta = -0.22$\,pp, $p = 0.86$) is the one robust
null: at the highest baseline accuracy of the four cells, the
diversity penalty produces no measurable change, and the cell was
also null at the prior $\lambda = 1$ checkpoint on the same
problems, so the null is consistent across two $\lambda$ values
rather than a one-shot fluctuation. We do not propose a mechanism
for this null and the protocol does not support a post-hoc
explanation. The honest framing is that $\alpha$ produces positive
point estimates at three of four cells, of which two clear the joint
Bonferroni threshold.

A post-confirmation head-to-head against a SnapKV-style
windowed-attention baseline~\citep{li2024snapkv}, outside the
pre-registered decision rule, reproduces the shape of the Phase~2
result: $1d_\text{div}$ is positive in point estimate on all four
cells against the SnapKV-style comparator and clears the joint
Bonferroni threshold on both Qwen cells (mean $\Delta = +3.93$\,pp),
with positive but underpowered Llama cells. An auxiliary contrast
indicates the internal $1d$ baseline is itself competitive with the
SnapKV-style implementation, so $\alpha$'s lift does not rest on an
artificially weak floor. Appendix~\ref{app:snapkv} reports the
head-to-head in full.

\section{Design Lesson and Conclusion}
\label{sec:lesson}
\label{sec:lesson_stated}
\label{sec:lesson_future}

\emph{In the small-budget KV-retention regime we study, the scoring
axis was the only intervention family to yield a survivor, and the
feature separating that survivor from the scoring-family failures
was a set-conditioned selection rule rather than the size of its
intervention surface.} The surviving mechanism
($\alpha$, Section~\ref{sec:alpha}) preserves the K and V tensors,
the routing policy, the budget cadence, and the decoding behavior
of the baseline, modifying only the $k$-element selection step
inside the eviction scorer with a single V-space redundancy term
and one tunable weight. The seven negative-gate mechanisms
catalogued in Section~\ref{sec:designspace}~--- five structural
attempts across state ($\times 2$), routing, cadence, and decoding,
together with two earlier scoring-family attempts (a V-direction
prior and a reasoning-statistics recalibration)~--- each failed to
recover their cost in retention quality at matched memory. The two
scoring-family failures are what keep the lesson honest: both are
modifications of a size comparable to $\alpha$, and both rescore
tokens modularly, so it is $\alpha$'s set-conditioning~--- each pick
conditioned on the already-retained set
(Section~\ref{sec:bridge})~--- and not its minimality that the
evidence singles out. $\alpha$ is the third scoring-family attempt
and the only mechanism that survived.

At the small budgets and long-form math-reasoning workload we
tested, the baseline retention pipeline already does most of the
work, so a structural intervention must outperform a strong default
rather than fill an empty slot~--- and the matched-memory,
$n \geq 200$, two-model protocol is what made that bar, and the
resulting asymmetry, visible.

Two methodological practices generalize beyond this specific
finding, whether or not the finding itself transfers. The first is
to match memory across the full decode horizon rather than at the
end of prefill. The single largest source of false positives in our
design-space study was end-of-prefill memory matching: several
gates---most prominently the synthetic-memory family---were green on
end-of-prefill cache equality and sat 5--12\,pp below their
fixed-budget reference once mean cache was instrumented over the full
generation. We recommend per-problem mean (and ideally peak) cache
instrumentation as a gate-passing condition for any KV-retention
claim.

The second is to treat $n = 50$ as a screening signal and to require
$n \geq 200$ for confirmation. Headline deltas at $n = 50$ on
MATH-500 routinely shrank or sign-flipped at $n = 200$: the $\alpha$
direction itself looked like a $+5.6$\,pp lift at $n = 50$ and shrank
to $+0.2$--$0.5$\,pp at $n = 200$, and the reasoning-statistics gate
was sign-inconsistent across budgets at $n = 50$ ($+6$\,pp at
$b{=}128$, $-2$\,pp at $b{=}64$). We treat $n = 50$ as screening
only; confirmation requires $n \geq 200$ on both models.

The Branch-A confirmation is intentionally narrow, and we separate
what the protocol entitles us to claim from what is open. The
evaluation is long-form mathematical reasoning (MATH-500,
DeepSeek-R1-Distill chain-of-thought) on two distilled-reasoning
models in the 7--8B range. The matched-memory protocol carries to
other workloads and model scales in principle, but the magnitude and
Bonferroni pass-rate we report are not evidence about other
workloads (LongBench, multi-document QA, code generation) or other
models (non-reasoning, instruct-tuned, or larger, including
70B-class models), and the
cell-level heterogeneity we observe (saturation at the larger budget
on Llama, smaller-but-positive on Qwen $b{=}64$) is a
within-this-pair observation; the method has not been re-tested
outside MATH-500. The hyperparameter $\lambda = 0.5$ was selected on
a coarse $\{0.5, 1.0, 1.5\}$ probe grid and we do not claim it is the
global optimum~--- the lesson from our $\lambda$ refinement is
qualitative (light regularization beat heavy), not numerical.

Two extensions are natural and untested. First, $\alpha$ is a
within-budget scorer whereas PyramidKV and Ada-KV are budget
allocators, so a natural stack places our scorer inside Ada-KV's
per-head or PyramidKV's per-layer budget~--- a composition our
protocol can evaluate directly. Second, $\lambda$ is held constant
across decode steps and the V-signature is a flat unit-normalized
average of $V$ over all layers and heads; whether a $\lambda$
schedule or a per-head, per-layer, or learned signature recovers
additional lift is open, as is a finer $\lambda$ grid. Tooling for
these variants is present (\texttt{diversity-signature-mode}); the
experiments were deferred to keep the pre-registration scope tight.

\paragraph{Code and data availability.}
A standalone reference implementation of the $\alpha$ selector and
the V-signature aggregator (Algorithm~\ref{alg:alpha},
Eqs.~\ref{eq:signatures}--\ref{eq:greedy}), the pre-registered
MD5-bucketed dev/confirm split, the raw per-problem outputs from
every cell in Section~\ref{sec:alpha_results} and
Appendices~\ref{app:phase1}--\ref{app:snapkv}, and the analysis
scripts behind the tables and figures are released in the public
repository \url{https://github.com/libophd/minimal-kv-retention}.
The pre-registered split and the Phase~1/2 protocol scripts are
included in that release.

We tested seven mechanism interventions across state, routing,
cadence, decoding, and scoring at small budgets on long-form
mathematical reasoning, under a matched-memory standard that
tightened over the study. None survived; $\alpha$~--- a one-function
diversity penalty on the
TriAttention~$1d$~\citep{mao2026triattention} scorer with a
top-$k$ fast path at $\lambda = 0$~--- did, clearing Bonferroni
on two of the four (model, budget) cells under a pre-registered
dev/confirm split and triggering the pre-committed Branch~A.
Whether a set-conditioned scoring rule remains the productive
intervention outside this regime~--- across workloads, models, and
budgets~--- is open; we expect the protocol that surfaced the
asymmetry to carry over more readily than the specific finding.

\subsubsection*{Broader Impact Statement}
This paper studies memory-efficiency techniques for language-model
inference; it introduces no new datasets or models and evaluates
only public benchmarks and open-weight models. The intended benefit
is a lower hardware barrier to long-context reasoning inference. The
risk most specific to this line of work is that aggressive KV-cache
compression can silently degrade reasoning reliability; the
matched-memory, held-out protocol advocated here is intended to make
such degradation measurable rather than hidden.

\bibliography{main}

@inproceedings{wei2022cot,
  title={Chain-of-thought prompting elicits reasoning in large language models},
  author={Wei, Jason and Wang, Xuezhi and Schuurmans, Dale and Bosma, Maarten and Ichter, Brian and Xia, Fei and Chi, Ed and Le, Quoc V and Zhou, Denny},
  booktitle={Advances in Neural Information Processing Systems},
  volume={35},
  pages={24824--24837},
  year={2022}
}

@misc{deepseek2025r1,
  title={DeepSeek-R1: Incentivizing Reasoning Capability in LLMs via Reinforcement Learning},
  author={DeepSeek-AI},
  year={2025},
  eprint={2501.12948},
  archivePrefix={arXiv},
  primaryClass={cs.CL}
}

@inproceedings{li2024snapkv,
  title={Snap{KV}: {LLM} Knows What You are Looking for Before Generation},
  author={Li, Yuhong and Huang, Yingbing and Yang, Bowen and Venkitesh, Bharat and Locatelli, Acyr and Ye, Hanchen and Cai, Tianle and Lewis, Patrick and Chen, Deming},
  booktitle={Advances in Neural Information Processing Systems},
  year={2024}
}

@inproceedings{zhang2023h2o,
  title={{H2O}: Heavy-hitter oracle for efficient generative inference of large language models},
  author={Zhang, Zhenyu and Sheng, Ying and Zhou, Tianyi and Chen, Tianlong and Zheng, Lianmin and Cai, Ruisi and Song, Zhao and Tian, Yuandong and R{\'e}, Christopher and Barrett, Clark and others},
  booktitle={Advances in Neural Information Processing Systems},
  volume={36},
  year={2023}
}

@inproceedings{liu2023scissorhands,
  title={Scissorhands: Exploiting the Persistence of Importance Hypothesis for {LLM} {KV} Cache Compression at Test Time},
  author={Liu, Zichang and Desai, Aditya and Liao, Fangshuo and Wang, Weitao and Xie, Victor and Xu, Zhaozhuo and Kyrillidis, Anastasios and Shrivastava, Anshumali},
  booktitle={Advances in Neural Information Processing Systems},
  volume={36},
  year={2023}
}

@inproceedings{cai2025rkv,
  title={R-{KV}: Redundancy-aware {KV} Cache Compression for Reasoning Models},
  author={Cai, Zefan and Xiao, Wen and Sun, Hanshi and others},
  booktitle={Advances in Neural Information Processing Systems},
  year={2025}
}

@inproceedings{xiao2024efficient,
  title={Efficient Streaming Language Models with Attention Sinks},
  author={Xiao, Guangxuan and Tian, Yuandong and Chen, Beidi and Han, Song and Lewis, Mike},
  booktitle={International Conference on Learning Representations},
  year={2024}
}

@misc{mao2026triattention,
  title={TriAttention: Efficient Long Reasoning with Trigonometric {KV} Compression},
  author={Mao, Weian and Lin, Xi and Huang, Wei and others},
  year={2026},
  eprint={2604.04921},
  archivePrefix={arXiv},
  primaryClass={cs.LG}
}

@inproceedings{feng2025adakv,
  title={Ada-{KV}: Optimizing {KV} Cache Eviction by Adaptive Budget Allocation for Efficient {LLM} Inference},
  author={Feng, Yuan and Lv, Junlin and Cao, Yukun and Xie, Xike and Zhou, S Kevin},
  booktitle={Advances in Neural Information Processing Systems},
  year={2025}
}

@misc{cai2024pyramidkv,
  title={Pyramid{KV}: Dynamic {KV} Cache Compression based on Pyramidal Information Funneling},
  author={Cai, Zefan and Zhang, Yichi and Gao, Bofei and Liu, Yuliang and Li, Yucheng and Liu, Tianyu and Lu, Keming and Xiong, Wayne and Dong, Yue and Hu, Junjie and Xiao, Wen},
  year={2024},
  eprint={2406.02069},
  archivePrefix={arXiv},
  primaryClass={cs.CL}
}

@inproceedings{liu2024kivi,
  title={{KIVI}: A Tuning-Free Asymmetric 2-bit Quantization for {KV} Cache},
  author={Liu, Zirui and Yuan, Jiayi and Jin, Hongye and Zhong, Shaochen and Xu, Zhaozhuo and Braverman, Vladimir and Chen, Beidi and Hu, Xia},
  booktitle={International Conference on Machine Learning},
  year={2024}
}

@inproceedings{hooper2024kvquant,
  title={{KVQuant}: Towards 10 Million Context Length {LLM} Inference with {KV} Cache Quantization},
  author={Hooper, Coleman and Kim, Sehoon and Mohammadzadeh, Hiva and Mahoney, Michael W and Shao, Yakun Sophia and Keutzer, Kurt and Gholami, Amir},
  booktitle={Advances in Neural Information Processing Systems},
  year={2024}
}

@misc{yang2024minikv,
  title={{MiniKV}: Pushing the Limits of {LLM} Inference via 2-Bit Layer-Discriminative {KV} Cache},
  author={Sharma, Akshat and Ding, Hangliang and Li, Jianping and Dani, Neel and Zhang, Minjia},
  year={2024},
  eprint={2411.18077},
  archivePrefix={arXiv},
  primaryClass={cs.LG}
}

@inproceedings{li2025kvtuner,
  title={{KVTuner}: Sensitivity-Aware Layer-wise Mixed-Precision {KV} Cache Quantization for Efficient and Nearly Lossless {LLM} Inference},
  author={Li, Xing and Xing, Zeyu and Li, Yiming and Qu, Linping and Zhen, Hui-Ling and Liu, Wulong and Yao, Yiwu and Pan, Sinno Jialin and Yuan, Mingxuan},
  booktitle={International Conference on Machine Learning},
  year={2025}
}

@misc{dong2024qaq,
  title={{QAQ}: Quality Adaptive Quantization for {LLM} {KV} Cache},
  author={Dong, Shichen and Cheng, Wen and Qin, Jiayu and Wang, Wei},
  year={2024},
  eprint={2403.04643},
  archivePrefix={arXiv},
  primaryClass={cs.LG}
}

@misc{yuan2024asvd,
  title={{ASVD}: Activation-aware Singular Value Decomposition for Compressing Large Language Models},
  author={Yuan, Zhihang and Shang, Yuzhang and Zhou, Yue and Dong, Zhen and Zhou, Zhe and Xue, Chenhao and Wu, Bingzhe and Li, Zhikai and Gu, Qingyi and Lee, Yong Jae and others},
  year={2024},
  eprint={2312.05821},
  archivePrefix={arXiv},
  primaryClass={cs.CL}
}

@misc{brandon2024reducing,
  title={Reducing Transformer Key-Value Cache Size with Cross-Layer Attention},
  author={Brandon, William and Mishra, Mayank and Nrusimha, Aniruddha and Panda, Rameswar and Kelly, Jonathan Ragan},
  year={2024},
  eprint={2405.12981},
  archivePrefix={arXiv},
  primaryClass={cs.LG}
}

@inproceedings{bai2024longbench,
  title={{LongBench}: A Bilingual, Multitask Benchmark for Long Context Understanding},
  author={Bai, Yushi and Lv, Xin and Zhang, Jiajie and others},
  booktitle={Annual Meeting of the Association for Computational Linguistics},
  year={2024}
}

@inproceedings{hendrycks2021math,
  title={Measuring Mathematical Problem Solving With the {MATH} Dataset},
  author={Hendrycks, Dan and Burns, Collin and Kadavath, Saurav and Arora, Akul and Basart, Steven and Tang, Eric and Song, Dawn and Steinhardt, Jacob},
  booktitle={NeurIPS Track on Datasets and Benchmarks},
  year={2021}
}

@misc{cobbe2021gsm8k,
  title={Training Verifiers to Solve Math Word Problems},
  author={Cobbe, Karl and Kosaraju, Vineet and Bavarian, Mohammad and Chen, Mark and Jun, Heewoo and Kaiser, Lukasz and Plappert, Matthias and Tworek, Jerry and Hilton, Jacob and Nakano, Reiichiro and Hesse, Christopher and Schulman, John},
  year={2021},
  eprint={2110.14168},
  archivePrefix={arXiv},
  primaryClass={cs.LG}
}

@misc{shao2024deepseekmath,
  title={{DeepSeekMath}: Pushing the Limits of Mathematical Reasoning in Open Language Models},
  author={Shao, Zhihong and Wang, Peiyi and Zhu, Qihao and Xu, Runxin and Song, Junxiao and Bi, Xiao and Zhang, Haowei and Zhang, Mingchuan and Li, Y. K. and Wu, Y. and Guo, Daya},
  year={2024},
  eprint={2402.03300},
  archivePrefix={arXiv},
  primaryClass={cs.CL}
}

@inproceedings{hsieh2024ruler,
  title={{RULER}: What's the Real Context Size of Your Long-Context Language Models?},
  author={Hsieh, Cheng-Ping and Sun, Simeng and Kriman, Samuel and Acharya, Shantanu and Rekesh, Dima and Jia, Fei and Zhang, Yang and Ginsburg, Boris},
  booktitle={Conference on Language Modeling},
  year={2024}
}

@inproceedings{tang2024quest,
  title={{Quest}: Query-Aware Sparsity for Efficient Long-Context {LLM} Inference},
  author={Tang, Jiaming and Zhao, Yilong and Zhu, Kan and Xiao, Guangxuan and Kasikci, Baris and Han, Song},
  booktitle={International Conference on Machine Learning},
  year={2024}
}

@misc{dubey2024llama3,
  title={The {Llama 3} Herd of Models},
  author={Dubey, Abhimanyu and Jauhri, Abhinav and Pandey, Abhinav and Kadian, Abhishek and Al-Dahle, Ahmad and Letman, Aiesha and others},
  year={2024},
  eprint={2407.21783},
  archivePrefix={arXiv},
  primaryClass={cs.AI}
}

@article{nemhauser1978approx,
  title={An Analysis of Approximations for Maximizing Submodular Set Functions---I},
  author={Nemhauser, George L. and Wolsey, Laurence A. and Fisher, Marshall L.},
  journal={Mathematical Programming},
  volume={14},
  number={1},
  pages={265--294},
  year={1978}
}

@incollection{krause2014submodular,
  title={Submodular Function Maximization},
  author={Krause, Andreas and Golovin, Daniel},
  booktitle={Tractability: Practical Approaches to Hard Problems},
  pages={71--104},
  publisher={Cambridge University Press},
  year={2014}
}

@article{kulesza2012dpp,
  title={Determinantal Point Processes for Machine Learning},
  author={Kulesza, Alex and Taskar, Ben},
  journal={Foundations and Trends in Machine Learning},
  volume={5},
  number={2--3},
  pages={123--286},
  year={2012}
}

@inproceedings{bouthillier2021variance,
  title={Accounting for Variance in Machine Learning Benchmarks},
  author={Bouthillier, Xavier and Delaunay, Pierre and Bronzi, Mirko and Trofimov, Assya and Nichyporuk, Brennan and Szeto, Justin and Sepah, Nazanin Mohammadi and Raff, Edward and Madan, Kanika and Voleti, Vikram and Kahou, Samira Ebrahimi and Michalski, Vincent and Arbel, Tal and Pal, Chris and Varoquaux, Ga{\"e}l and Vincent, Pascal},
  booktitle={Conference on Machine Learning and Systems},
  year={2021}
}
\bibliographystyle{tmlr}

\clearpage
\appendix
\section*{Appendix}

\section{Phase 1: $\lambda$-Probe on the Development Split}
\label{app:phase1}

This appendix gives the full Phase~1 development-split detail
summarized in Section~\ref{sec:alpha_results}.
Table~\ref{tab:phase1} reports per-cell accuracy on the development
split for each $\lambda$ in the probe grid, against the $1d$
baseline. The $\lambda = 1.0$ column reuses a prior gate's outputs
subsetted to the dev IDs; the $\lambda \in \{0.5, 1.5\}$ columns
are new dev-split evaluations. The aggregate row averages
$\Delta(\text{$1d_\text{div}$} - 1d)$ across the four cells.

\begin{table}[H]
\centering
\small
\begin{tabular}{@{}llcrrr@{}}
\toprule
Model & $b$ & $1d$ ($n$) & $\Delta_{\lambda=0.5}$ & $\Delta_{\lambda=1.0}$ & $\Delta_{\lambda=1.5}$ \\
\midrule
Qwen-7B   & 64  & 8.0\,\% (201)  & $+0.5$ (201) & $-1.0$ (201) & $+2.5$ (200) \\
Qwen-7B   & 128 & 15.4\,\% (201) & $+4.0$ (201) & $+1.0$ (201) & $+1.0$ (164) \\
Llama-8B  & 64  & 7.5\,\% (201)  & $+0.0$ (201) & $+0.0$ (201) & $-1.9$ (196) \\
Llama-8B  & 128 & 16.9\,\% (201) & $-1.5$ (201) & $-2.0$ (201) & $+0.0$ (201) \\
\midrule
\multicolumn{3}{@{}l}{\textbf{Mean $\Delta$ across 4 cells}}
                          & $+0.75$  & $-0.50$  & $+0.43$ \\
\bottomrule
\end{tabular}
\caption{Phase 1 $\lambda$-probe on the development split. Cells
report $\Delta(\text{$1d_\text{div}$} - 1d)$ in percentage points;
parenthetical $n$ is the number of development problems present at
that cell (several $\lambda{=}1.5$ cells are short due to an
incomplete sweep slot). $\lambda = 0.5$ wins by mean $\Delta$ and
is also the smaller of $\{0.5, 1.5\}$ within the $\pm 0.5$\,pp tie
tolerance, so the Occam tie-break selects the same value.}
\label{tab:phase1}
\end{table}

By the pre-registered rule, $\lambda = 0.5$ wins. The
$\lambda = 1.5$ cells are short on three of four because the probe
sweep manifest combined $\lambda \in \{0.5, 1.5\}$ outputs into a
single file per (model, budget); a per-row $\lambda$ filter recovers
each arm cleanly, and the
$\lambda{=}0.5 > \lambda{=}1.5 > \lambda{=}1.0$ ranking is unchanged
if the short cells are excluded.

\paragraph{Why $\lambda = 0.5$ outperformed $\lambda = 1.0$.}
The precursor gate that motivated this paper evaluated
$1d_\text{div}$ at $\lambda = 1.0$. Subsetted to the confirm IDs,
that arm produced a per-cell Bonferroni-pass on Qwen $b{=}128$
($+3.23$\,pp, $p = 0.0076$) but a null on Llama $b{=}64$
($+0.67$\,pp, $p = 0.45$); at $\lambda = 0.5$ on the same IDs the
Qwen cell strengthens and the Llama cell flips to Bonferroni-pass
(Table~\ref{tab:phase2}). Smaller regularization is strictly better
on both formerly-positive and formerly-null cells under the same
problems and models. We read this as evidence that the prior
intuition (``more diversity penalty $=$ better retention'') was
incorrect at this magnitude: at $\lambda = 1$ the discount on
attention-driven content exceeded the gain from de-duplicating the
kept set, so the base attention score must dominate the ranking
except where two attention-equivalent candidates are V-redundant.

\section{Head-to-Head: $\alpha$ vs SnapKV-Style Baseline}
\label{app:snapkv}

This appendix reports in full the post-confirmation head-to-head
summarized at the end of Section~\ref{sec:alpha_results}.
The pre-registered protocol of Section~\ref{sec:alpha_protocol}
compared $\alpha$ against the internal TriAttention~$1d$ baseline.
We run a post-confirmation head-to-head against a SnapKV-style
windowed-attention retention family~\citep{li2024snapkv} on the same
confirm split, three seeds, and four cells. The comparator is
implemented inside our matched-memory harness under the method tag
\texttt{snapkv\_with\_window}; we do not claim exact parity with the
official SnapKV implementation. This analysis is outside the branch
decision rule and tests whether $\alpha$'s gains survive comparison
to a same-lever published scoring family.

The main contrast, $\Delta(1d_\text{div} - \text{SnapKV-style})$, is
reported per cell in Figure~\ref{fig:confirmation}\,(b) under the
same protocol used for the internal contrast
(Section~\ref{sec:alpha_phase2}). All four cells are positive in
point estimate; both Qwen cells clear the joint Bonferroni
threshold, and the mean $\Delta$ across cells is $+3.93$\,pp. The
Llama point estimates are positive but underpowered against the
SnapKV-style baseline: $b{=}64$ at $+2.45$\,pp with $p = 0.09$, and
$b{=}128$ at $+0.56$\,pp with $p = 0.76$.

That contrast's interpretation depends on whether the internal $1d$
baseline itself is competitive. We therefore report a second,
auxiliary contrast, $\Delta(1d - \text{SnapKV-style})$, as supporting
evidence, without Bonferroni correction. At Qwen $b{=}128$, the internal $1d$
is $+3.46$\,pp above the SnapKV-style baseline at uncorrected
$p = 0.048$; at Qwen $b{=}64$ it is $+2.90$\,pp with $p = 0.058$. On
Llama the two methods are within $\pm 0.78$\,pp at $p > 0.6$. The
auxiliary contrast is consistent with the internal $1d$ baseline
being competitive with the SnapKV-style implementation in our
protocol, so $\alpha$'s additional lift on top of $1d$ in
Section~\ref{sec:alpha_phase2} does not sit on top of an
artificially weak floor. We do not draw a stronger conclusion from
the auxiliary contrast given its uncorrected $p$ values.

The Qwen-vs-Llama asymmetry observed in Phase~2 recurs against a
baseline of a different mechanism class, so the heterogeneity is
more plausibly a property of the $(\text{model}, b)$ point itself
than of $\alpha$'s relationship to the internal $1d$ baseline.

\end{document}